\documentclass[10pt,twocolumn,cls]{IEEEtran}
\usepackage{amssymb}
\usepackage{graphicx,subfigure}
\usepackage{amsmath}
\usepackage{psfig,epsfig,algorithm,amsmath,amssymb,color,subfigure,url}
\usepackage{algorithmic}
\usepackage[center]{caption}

\newcommand{\stable}{\text{stable}}
\newcommand{\detection}{\text{detection}}
\newcommand{\rotation}{\text{rotation}}
\newcommand{\status}{\text{status}}
\newcommand{\del}{\text{del}}
\newcommand{\new}{\text{new}}
\newcommand{\pred}{\text{pred}}

\author{Chenlu Qiu, Namrata Vaswani \\
Dept. of Electrical and Computer Engineering, Iowa State University\\
Ames, IA, 50010\\
\{chenlu, namrata\}@iastate.edu
\thanks{This research was partially supported by NSF grants ECCS-0725849 and CCF-0917015.}
}
\begin{document}

\title{Real-time Robust Principal Components' Pursuit}
\maketitle

\begin{abstract}
In the recent work of Candes et al, the problem of recovering low rank matrix corrupted by i.i.d. sparse outliers is studied and a very elegant solution, principal component pursuit, is proposed. It is motivated as a tool for video surveillance applications with the background image sequence forming the low rank part and the moving objects/persons/abnormalities forming the sparse part. Each image frame is treated as a column vector of the data matrix made up of a low rank matrix and a sparse corruption matrix. Principal component pursuit solves the problem under the assumptions that the singular vectors of the low rank matrix are spread out and the sparsity pattern of the sparse matrix is uniformly random. However, in practice, usually the sparsity pattern and the signal values of the sparse part (moving persons/objects) change in a correlated fashion over time, for e.g., the object moves slowly and/or with roughly constant velocity. This will often result in a low rank sparse matrix.

For video surveillance applications, it would be much more useful to have a real-time solution. In this work, we study the online version of the above problem and propose a solution that automatically handles correlated sparse outliers. In fact we also discuss how we can potentially use the correlation to our advantage in future work. The key idea of this work is as follows. Given an initial estimate of the principal directions of the low rank part, we causally keep estimating the sparse part at each time by solving a noisy compressive sensing type problem. The principal directions of the low rank part are updated every-so-often. In between two update times, if new Principal Components' directions appear, the ``noise" seen by the Compressive Sensing step may increase. This problem is solved, in part, by utilizing the time correlation model of the low rank part. We call the proposed solution ``Real-time Robust Principal Components' Pursuit". It still requires the singular vectors of the low rank part to be spread out, but it does not require i.i.d.-ness of either the sparse part or the low rank part.

\end{abstract}

\section{Introduction}
Principal Components' Analysis (PCA) tries to find the ``principal components' space" with the smallest dimension that spans a given dataset. In practice, data is noisy and in this case PCA finds the smallest subspace to represent the dataset with a given mean squared error (MSE) tolerance.

Given a low rank data matrix $M \in \mathbb{R}^{m \times n}$ (each column of $M$ is one data vector), PCA finds its principal components (PCs) as the left singular vectors of $M$ that have nonzero singular values. This is the same as first estimating the data covariance as $(1/n)MM^T$, computing its eigenvalue decomposition (EVD) and retaining eigenvectors corresponding to nonzero eigenvalues. When data is noisy, this is replaced by arranging the eigenvectors in decreasing order of eigenvalues, and retaining the smallest number of eigenvectors so that the sum of the remaining eigenvalues (which is equal to the residual MSE) is less than the MSE tolerance.

When the noise is small, the above approach works well. However, covariance matrix estimation, and hence the corresponding EVD, are sensitive to even a few large outliers in the data. Unfortunately, in practice these do occur, e.g. when trying to compute the principal components' subspace for a video sequence, parts of it may get occluded by other moving objects. There has been a large amount of work in literature on ``Robust PCA", e.g. \cite{rpca_1,rpca_caramanis,Roweis98emalgorithms,rpca_2,sequentialSVD,Li03anintegrated,ipca_5,ipca_weightedand,Torre01robustprincipal}, most of which either assumes the locations of the missing/corruped data points are known \cite{Roweis98emalgorithms}, which is not a practical assumption, or (ii) first tries to detect the corrupted pixels and then either fills in the corrupted location using some heuristics or (iii) often just removes the entire outlier vector.
In a series of recent works \cite{rpca,rpca_Chandrasekaran,error_correction_PCP}, a very elegant solution to this problem was provided that treats the outlier as a sparse vector. In \cite{rpca}, the data matrix $M$ consists of a low rank matrix that is corrupted by sparse outliers, i.e.
\begin{equation*}
M = L + S
\end{equation*}
where $L$ is a low rank matrix having a singular value decomposition (SVD) $L \overset{SVD}{=} U D V^T$ and $S$ is sparse and can have arbitrary large magnitude. Let $\|L\|_*$ denotes the nuclear norm of $L$, i.e., the sum of singular values of $L$. It is shown in \cite{rpca} that $L$ and $S$ can be recovered with high probability by solving a convex optimization problem, named as Principal Component Pursuit (PCP),
\begin{eqnarray}
&&\underset{L,S} {\min} \ \ \ \ \ \ \ \ \ \|L\|_* + \lambda\|S\|_1 \label{PCP}\\
&&\text{subject to}  \ \ \ L +S = M \nonumber
\end{eqnarray}
provided the singular vectors of $L$ are spread out (not sparse), the support and signs of $S$ are uniformly random (thus not low rank), the rank of $L$ and the fraction of corrupted entries in $S$ are both sufficient small. A more recent work, \cite{error_correction_PCP}, extends the result of \cite{rpca} showing that, with a proper weighting parameter $\lambda$, PCP can recover $L$ and $S$ with high probability even if the size of support set of $S$ is large, as long as the rank of $L$ is small enough. But it requires that $S$ has random support and random signs.

PCP \cite{rpca} is motivated as a tool for surveillance applications, with the background variations approximately lying in a low dimension subspace, and the sparse part being the ``moving persons" or ``abnormalities" to be detected. It is an offline method which treats each image frame as a column vector of the data matrix $M$. While this is a very elegant and novel idea, there are certain limitations.
\begin{itemize}
\item [1)] In surveillance, it would be more useful to obtain the estimates of the sparse part on-the-fly rather than offline.
\item [2)] The sparsity pattern (support and signs) of the sparse part may change slowly or in a correlated fashion, which may result in a low rank sparse matrix. In this case, PCP assumption will not get satisfied and as a result it will not work, e.g. see Fig.\ref{plot1}.
\item [3)] The principal directions (set of eigenvectors corresponding to nonzero eigenvalues) can change over time. So the rank of the matrix $L$ will keep increasing over time thus making PCP impossible to do after sometime.
\end{itemize}
This last issue may get resolved by not using all frames of $M$, but only the latest image frames. But the first two issues still remain.

In this paper, we propose an online approach to solve this problem. Our goal is to causally keep estimating the sparse part $S_t$ at each time, and to keep updating the principal directions every-so-often. The $t$-th column of $M$, $M_t$, is the data acquired at time $t$. It can be split as
\begin{equation}
M_t = L_t + S_t = [U \ I] \left[
                            \begin{array}{c}
                              x_t \\
                              S_t \\
                            \end{array}
                          \right] \label{model_JP}
\end{equation}
where $x_t := U^T L_t$ and the matrix $U$ is an \emph{unknown} $m \times m$ orthonormal matrix. The support of the vector $S_t$ changes slowly over time.
Given an initial estimate of $P_t$ := $(U)_{N_t}$, denoted $\hat{P}_t$, we solve for the sparse vector $S_t$ by first finding the orthogonal complement matrix $\hat{P}_{t,\perp}$ and then using the projection of $M_t$ onto $\hat{P}_{t,\perp}$, denoted by $y_t$,
\begin{equation*}
y_t := \hat{P}_{t,\perp}^T M_t =  \hat{P}_{t,\perp}^T L_t +  \hat{P}_{t,\perp}^T S_t.
\end{equation*}
to solve for $S_t$. Notice that if $\hat{P}_t \approx P_t$ the first term will be close to zero and can be treated as ``noise". When $\hat{P}_t \neq P_t$ (new directions added), the ``noise" can be reduced by using the time correlation model on $L_t$. Furthermore, recent estimates of $L_t:=M_t - S_t$ are stored and used to periodically update $P_t$ as described in Sec. \ref{update_P}. There are also some limitations of our method.
\begin{itemize}
\item [1)] We need an approximately accurate initial estimate of the PCs' basis, $\hat{P}_0$, which is easy to get using training data without sparse corruptions.
\item [2)] The orthogonal complement $\hat{P}_{t,\perp}$ needs to satisfy some conditions for Compressive Sensing to succeed.
\item [3)] An appropriate choice of constraint parameter $\epsilon$ is needed for estimating $S_t$.
\end{itemize}

The above idea is somewhat related to that of \cite{decodinglp} in that both try to cancel the ``message" signal and only solve for the sparse ``error" signal, but with the big difference that in \cite{decodinglp}, $P_t$ is \emph{known}. Other related work which also uses $P_t$ known is \cite{rpca_regression, rpca_regression_sparse}. However in our problem $P_t$ is \emph{unknown }and can change with time.
Out method requires the columns of $\hat{P}_{t,\perp}$ be spread out (not sparse), but it does not require $S_t$ to have independent nonzero entries. In fact, we can utilize the correlated support change of $S_t$ over time, $t$, to our advantage in future work. Since we update the principal directions on-the-fly, the dimension of the principal subspace remains bounded.

A model similar to (\ref{model_JP}) but for a static problem and with $U$ being a \emph{known} matrix, was introduced in \cite{error_correction_PCP_l1, Laska_exactsignal}. A method, termed as pursuit of justice (PJ), is introduced to solve for the sparse vector $u = [x_t, S_t]^T$ which solving the following $\ell_1$ minimization problem
\begin{eqnarray}
&&\underset{u} {\min} \ \ \ \ \ \ \ \ \ \|u\|_1  \label{PJ}\\
&&\text{subject to}  \ \ \ M_t = A u \nonumber
\end{eqnarray}
where $A := [U \ I]$.
Notice that in our problem $U$ is \emph{unknown}, and thus we cannot use sparse reconstruction techniques to find $x_t$. Given an estimate $\hat{P}_t$, the above can be modified to $A = [ \hat{P} \ \hat{P}_{t,\perp} \ I]$. However, this does not work as shown in Fig. \ref{plot2}.


\subsection{Notations}
The set operations $\cup$, $\cap$ and $\setminus$ have the usual meanings. For any set $T \subset \{1 , \cdots m\}$, $T^c$ denotes the complement set of $T$, i.e., $T^c : = \{1 , \cdots m\} \setminus T$.

For a non diagonal matrix $A$, we let $A_i$ denote the $i$th column of $A$ and we let $A_T$ denote a matrix composed of the columns of $A$ indexed by $T$. For two set $T_1$ and $T_2$, we let $A_{T_1,T_2}$ denote a submatrix of $A$ consisting of the rows indexed by $T_1$ and columns indexed by $T_2$.
For a diagonal matrix $Q$, $Q_T$ denotes a submatrix of $Q$ consisting of the rows and columns indexed by $T$. In other words, $Q_T$ is a diagonal matrix with $(Q_T)_{j,j} = (Q)_{T_j,T_j}$.

For vector $v$, $v_i$ denotes the $i$th entry of $v$ and $v_T$ denotes a vector consisting of the entries of $v$ indexed by $T$. $\|v\|_k$ denotes the $\ell_k$ norm of $v$. The support of $v$, $\text{supp}(v)$, is the set of indices at which $v$ has nonzero value, $\text{supp}(v): = \{i:\ v_i \neq 0\}$.

We use $\emptyset$ to denote an empty set or an empty matrix.

\section{Problem definition and Signal Model}

The $t$th column of $M$, $M_t \in \mathbb{R}^{m \times 1}$, is the data at time $t$ which can be split as
\begin{eqnarray*}
M_t &=& L_t + S_t \\
L_t &=& U x_t = P_t a_t
\end{eqnarray*}
where $x_t : = U^T L_t$ and $S_t$ are sparse vectors with slowly changing support $N_t:= \text{supp}(x_t)$ and $T_t := \text{supp}(S_t)$, respectively. $N_t$ is modeled as being piecewise constant with time. The vector $a_t  := (x_t)_{N_t} $ is the none-zero part of $x_t$. The principal components' basis at each time $t$, $P_t := U_{N_t}$, is a submatrix of $U$ whose columns span the principal components' subspace at time $t$. It is unknown and can change over time.

Since the matrix $U$ does not change with time (in this work), the only way $P_t$ changes is when the set $N_t$ changes. This happens every $d$ frames.
We assume that $x_t$ and hence $L_t = U x_t$ follows a {\em piecewise stationary model with nonstationary transients when switching pieces}. For every $d$ frames, there are some supporting indices get added or deleted from $N_t$. Specifically when an element $i$ gets added into the support, it gets added with an initial small variance $\theta \sigma_i^2$ (with $0<\theta < 1$) and then at future times follows a first order autoregressive (AR-1) model with AR parameter $f$ and stable variance $\sigma_i^2$. Recall that an AR-1 model is asymptotically stationary. Thus, after the initial transient period, $x_t$ is stationary until the next support change time.
Before an element $i$ gets deleted, it starts decaying as $(x_t)_i = f_d (x_{t-1})_i$, with $0<f_d < f < 1$, and soon decays to zero.

\subsection{Mathematical description of signal model for $x_t$ (and hence for $L_t$)}\label{model_L}
The support set of $x_t$, $N_t$, is a union of three disjoint sets $\Delta_t$, $D_t$, and $E_t$, i.e., $N_t = \Delta_t \cup D_t \cup E_t$. The addition set $\Delta_t := N_t \setminus N_{t-1}$ is the set of indices for the new appearing eigenvectors $(U)_{\Delta_t}$. The set $D_t \subset (N_t \cap N_{t-1})$ is the set of indices of those eigenvectors whose eigenvalues are decreasing at time $t$. The set $E_t := N_t \cap N_{t-1} \setminus D_t$ is the set of indices for existing eigenvectors with non-decreasing eigenvalues. The sets $D_t$ and $\Delta_t$ can be empty. For any time $\tau$ with ``decreasing" set $D_{\tau}$, we assume that $D_{\tau}$ will not get added to $N_t$ for any $t >\tau$.

Let $\Sigma = \text{diag}(\sigma_i^2), \ i =1,\cdots,m, $ be a diagonal matrix with non-increasing positive diagonal elements, i.e. $\sigma_i^2$ satisfying $\sigma_i^2 \geq \sigma_{i+1}^2$. We model $x_t$ as
\begin{eqnarray}
&&x_0 = 0, \ N_0 = \emptyset \nonumber \\
&& x_t = F_t x_{t-1} + \nu_t, \ \nu_t \overset{i.i.d.}{\sim} \mathcal{N}(0,Q_t) \label{signal_model}
\end{eqnarray}
where $F_t$ and $Q_t$ are two diagonal matrices defined as below
\begin{eqnarray*}
&& (F_t)_{\Delta_t} = 0, \ (Q_t)_{\Delta_t} = \theta (\Sigma)_{\Delta_t}, \\
&&(F_t)_{E_t} = f I, \ (Q_t)_{E_t} = (1-f^2) (\Sigma)_{E_t}, \\
&&(F_t)_{D_t} = f_d I, \ (Q_t)_{D_t} = 0, \\
&&(F_t)_{N_t^c} = 0,\ (Q_t)_{N_t^c} = 0.
\end{eqnarray*}
where $f$, $f_d$, and $\theta$ are scalars satisfying $0< f_d <f <1$ and $0 < \theta <1$.

From the model on $x_t$, we notice the following:
\begin{itemize}

\item[a)] At time $t = \tau$, $(x_{\tau})_{\Delta_{\tau}}$ starts with
\begin{equation*}
(x_{\tau})_{\Delta_{\tau}} \sim \mathcal{N} (0, \theta (\Sigma)_{\Delta_{\tau}}).
\end{equation*}
Small $\theta$ ensures that new directions gets added at a small value and increase slowly.
$(x_{\tau})_{D_{\tau}}$ decays as \begin{equation*}
(x_{\tau})_{D_{\tau}} = f_d (x_{\tau-1})_{D_{\tau}}
\end{equation*}
$(x_{\tau})_{E_{\tau}}$ follows an AR-1 model with parameter $f$:
\begin{equation*}
(x_{\tau})_{E_{\tau}} = f (x_{\tau-1})_{E_{\tau}} + (v_{\tau})_{E_{\tau}}
\end{equation*}

\item[b)] At time $t>\tau$, if $\Delta_{\tau}$ is not removed from the support set, the variance of $(x_t)_{\Delta_{\tau}}$ gradually increases as
\begin{equation*}
(x_t)_i \sim \mathcal{N}( 0, (1 -(1-\theta)f^{2(t-\tau)})  \Sigma_{i,i}), \ \ i \in \Delta_{\tau}
\end{equation*}
Eventually, the variance of $(x_t)_{\Delta_{\tau}}$ converges to $(\Sigma)_{\Delta_{\tau}}$. For example, with $f = 0.9$ and $\theta = 0.4$, the variance of $(x_t)_{\Delta_{\tau}}$ gets to $0.9 (\Sigma)_{\Delta_{\tau}}$ in $18$ frames.

\item[c)] At time $t>\tau$, the variance of $(x_t)_{D_{\tau}}$ decays as
\begin{equation*}
(x_t)_{D_{\tau}} \sim \mathcal{N} (0, f_d^{2(t-\tau)} (\Sigma)_{D_{\tau}})
\end{equation*}
Eventually, $(x_t)_{D_{\tau}}$ decays to zero. For example, with $f_d =0.1$, the variance of $(x_t)_{D_{\tau}}$ decrease to $ 0.0001 (\Sigma)_{D_{\tau}}$ in $2$ frames.

\end{itemize}

\subsection{Model for $S_t$}\label{model_S}

Recall that in video surveillance applications, the data matrix $M$ is obtained by stacking each image frame as a column vector, whose low rank component $L$ corresponds to background variation lying in a low rank subspace and sparse component $S$ captures the moving objects in the foreground. In this work, we use a simple model for the sparse component $S$ modeling the activity of the moving objects as described below.

We assume that, in each image frame, there are $k$ ($k \geq1 $) objects in the foreground. Each object occupies a $3 \times 3$ pixel block which has nonzero pixel values. All other pixels in the foreground have zero values.
Let $CG_t^i$ denote the coordinate of the center of gravity of the $i$th object at time $t$, $i = 1, \cdots, k$.
For the next image frame, each $CG_t^i$ can either be static with probability $p$ or move one step to the left/right/top/bottom with probability $(1-p)/4$ each, i.e., for $ i = 1, \cdots, k$,
\[
CG_t^i= \left\{ \begin{array}{ll}
CG_{t-1}^i & \mbox{with probability $p$} \\
CG_{t-1}^i + (1,0), & \mbox {with probabilty $(1-p)/4$ }\\
CG_{t-1}^i + (-1,0), & \mbox {with probability $(1-p)/4$}\\
CG_{t-1}^i + (0,1), & \mbox {with probability $(1-p)/4$ }\\
CG_{t-1}^i + (0,-1), & \mbox {with probability $(1-p)/4$}\\
\end{array}
\right. \\
\]
with $p = 0.8$. The pixels in each block move accordingly. Except if the objects move very fast or if they are very small, there will be overlap between their regions from frame to frame. We then stack the resulting foreground image frame as columns of $S$. Clearly, the support of $S_t$, $t$th column of $S$, is time correlated and the signs of these nonzero entries are fixed. This is quite different from \cite{rpca} and \cite{error_correction_PCP} where random support and random signs are assumed on the sparse part $S$.

\section{Real-time Robust PCP}
\begin{figure*}
\centerline{
\psfig{file = 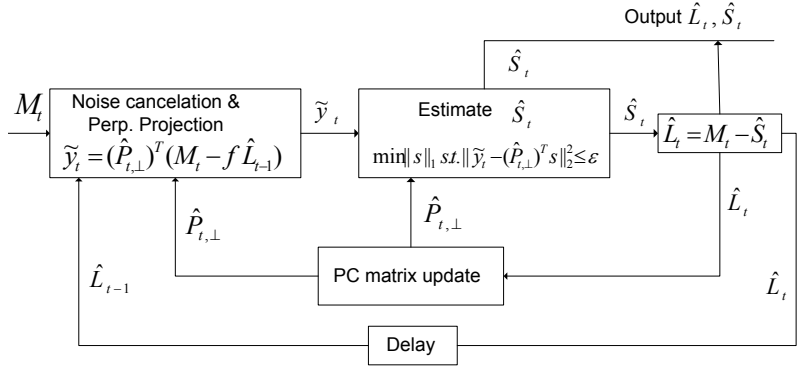, height = 5cm}
}
\caption{Real-time Robust PCP}
\label{chart1}
\end{figure*}

An overview of our method, real-time robust PCP (RR-PCP), is shown in Fig.\ref{chart1}. We first discuss the approach to recursively reconstruct the sparse component $S_t$. Next, we discussed the way we track the changes of the principal directions. Finally, a complete algorithm is given in Algorithm \ref{algorpcp}.

\subsection{RR-PCP: recursively reconstruction of the sparse part $S_t$}\label{main_idea}
Using $\hat{P}_t$, which is an estimate of principal components $P_t$ at time $t$, we can rewrite $L_t$ and $M_t$ as
\begin{eqnarray}
L_t &=& \hat{P}_t \alpha_t + \hat{P}_{t, \perp} \beta_t \nonumber \\
M_t &=& \hat{P}_t \alpha_t +  \hat{P}_{t, \perp} \beta_t  + S_t \nonumber
\end{eqnarray}
where $\hat{P}_{t,\perp}$ is an orthogonal complement of $\hat{P}_t$; $\alpha_t := \hat{P}_t^T L_t$ is the projection of $L_t$ onto the subspace spanned by $\hat{P}_t$; and $\beta_t := \hat{P}_{t, \perp}^T L_t$ is the projection of $L_t$ onto the subspace spanned by $\hat{P}_{t,\perp}$.
Notice that $\hat{P}_t$ is an estimate of $P_t$. It is either just a slight rotation of $P_t$ with $\text{span}(P_t) =\text{span}(\hat{P}_t)$ or there may be some missing and extra principal directions. The column vectors of $\hat{P}_{t,\perp}$ are the eigenvectors spanning the null space of $\hat{P}_t^T$. The orthogonal complement $\hat{P}_{t,\perp}$ is not unique.

Let
\begin{equation}
y_t := \hat{P}_{t,\perp}^T M_t = \hat{P}_{t,\perp}^T S_t + \beta_t \label{split}
\end{equation}
If there is no missing principal direction, i.e., $\text{span}(P_t) \subseteq \text{span}(\hat{P}_t)$, $\beta_t = 0$. If there are missing principal directions, $\text{span}(P_t) \nsubseteqq \text{span}(\hat{P}_t)$ and $\beta_t\neq 0$. In this case, $\beta_t$ in (\ref{split}) is the ``noise" resulting from the estimation error of current principal directions.
This now becomes a noisy sparse reconstruction problem, with ``noise" $\beta_t$. When $\|\beta_t \|_2^2$ is not very large, we can causally recover $S_t$ by solving
\begin{equation}
\min_{s} \|s\|_1 \ s.t. \ \| \hat{P}_{t,\perp}^T (M_t - s)\|_2^2 \le \epsilon  \label{cs}
\end{equation}
and hence estimate $L_t$ as
\begin{equation*}
\hat{L}_t= M_t -\hat{S}_t
\end{equation*}
where $\epsilon$ is a parameter with some small positive value. 

For the case of missing principal directions, $\text{span}(P_t) \nsubseteqq \text{span}(\hat{P}_t)$. Let $\mathcal{S}_{t,\text{miss}} : = \text{span}(P_t) \setminus \text{span}(\hat{P}_t)$ denote the ``missing" subspace and let $P_{t,\text{miss}}$ be its orthonormal basis matrix. Thus $\text{span}(P_{t,\text{miss}}) = \mathcal{S}_{t,\text{miss}}$ and
\begin{eqnarray}
\text{span}(P_t) &=& \text{span}(\hat{P}_t) \oplus \text{span}(P_{t,\text{miss}}) \\
\text{span}(\hat{P}_{t,\perp}) &=& \text{span}(P_{t,\perp}) \oplus \text{span}(P_{t,\text{miss}})
\end{eqnarray}
Therefore,
\begin{equation}
\beta_t = \hat{P}_{t,\perp}^T L_t = \hat{P}_{t,\perp}^T P_{t,\text{miss}} P_{t,\text{miss}}^T L_t
\end{equation}
with $P_{t,\text{miss}}^T L_t $ being the projection of $L_t$ onto the subspace $\mathcal{S}_{t,\text{miss}}$. If $P_{t,\text{miss}}^T L_t $ starts with small values, $\| \beta_t \|_2^2$ shall be small and it can increase over time. When $\|\beta_t\|_2^2$ is getting too large, (\ref{cs}) may give incorrect estimate $\hat{S}_t$. Thus, we need to update $\hat{P}_t$ and get those missing directions detected.

\subsection{Canceling the ``noise" using the time correlation of $x_t$}

It is expensive to update $\hat{P}_t$ and $\hat{P}_{t,\perp}$ very frequently, especially for some real-time applications.
But notice that we can cancel out some of $\beta_t$ by using the model on $x_t$ from Sec.\ref{model_L}. We modify (\ref{cs}) as
\begin{equation}
\min_{s} \|s\|_1 \ s.t. \ \| \hat{P}_{t,\perp}^T (M_t - s - f \hat{L}_{t-1})\|_2^2 \le \epsilon \label{csres}
\end{equation}
Let $\hat{\beta}_{t-1} := \hat{P}_{t, \perp}^T \hat{L}_{t-1}$. Note that in (\ref{csres}), the ``noise" is $\beta_t - f \hat{\beta}_{t-1}$ while in (\ref{cs}), the ``noise" is $\beta_t$.

Next, we discuss an example showing that the ``noise" $\beta_t - f \hat{\beta}_{t-1}$ in (\ref{csres}) is smaller than the ``noise" $\beta_t$ in (\ref{cs}).

Suppose at time $t=\tau-1$, we have an exact estimate of all principal directions, $\hat{P}_{\tau-1} = P_{\tau-1}$.
At time $t=\tau$, a support change occurs with $N_{\tau} = N_{\tau-1} \cup \Delta_{\tau}$ and $D_{\tau} = \emptyset$.
The principal directions at time $t=\tau$ are $P_{\tau} = [P_{\tau-1}, \ P_{\Delta_{\tau}} ]$ where $P_{\Delta_{\tau}} = U_{\Delta_{\tau}}$ are the new added principal directions.
However, this change is unknown to us and we just use $\hat{P}_{\tau} = \hat{P}_{\tau-1} = P_{\tau-1}$. Therefore,
\begin{equation*}
\text{span}(P_{\tau}) = \text{span}(P_{\tau-1}) \oplus \text{span}(P_{\Delta_{\tau}})
\end{equation*}
Thus, at time $t\geq \tau$,
\begin{eqnarray}
\beta_t &=& \hat{P}_{\tau,\perp}^T L_t = \hat{P}_{\tau,\perp}^T (P_{\tau-1} P_{\tau-1}^T L_t + P_{t,\Delta_{\tau}}P_{t,\Delta_{\tau}}^T L_t) \nonumber \\
&=& \hat{P}_{\tau,\perp}^T P_{t,\Delta_{\tau}}P_{t,\Delta_{\tau}}^T U x_t
\end{eqnarray}
Assuming $L_{t-1}$ is correctly recovered, i.e., $\hat{L}_{t-1} = L_{t-1}$, so $\hat{\beta}_{t-1} = \beta_{t-1}$, then
\begin{eqnarray*}
\beta_t  - f \hat{\beta}_{t-1} &=& \hat{P}_{\tau,\perp}^T P_{t,\Delta{\tau}}P_{t,\Delta_{\tau}}^T U  \left[ \begin{array}{c} (\nu_t)_{\Delta_{\tau}} \\  (\nu_t)_{E_{\tau}}  \\ 0 \\ \end{array}  \right] \\
&=& \hat{P}_{\tau,\perp}^T P_{t,\Delta{\tau}} (\nu_t)_{\Delta_{\tau}}
\end{eqnarray*}

Let $B = (\hat{P}_{\tau,\perp}^T P_{t,\Delta{\tau}})^T (\hat{P}_{\tau,\perp}^T P_{t,\Delta{\tau}})$, then at time $t \geq \tau$,
\begin{eqnarray}
\mathbb{E}(\|\beta_t\|_2^2) &=& \underset {i \in \Delta_{\tau}}{\Sigma} B_{i,i} (1- (1-\theta)f^{2(t-\tau)}) \sigma_i^2 \nonumber \\
\mathbb{E}(\|\beta_t - f \hat{\beta}_{t-1}\|_2^2)&=&
\underset {i \in \Delta_{\tau}}{\Sigma} B_{i,i} (1-f^2) \sigma_i^2 \label{eqn_beta}
\end{eqnarray}

Clearly, $\mathbb{E}(\|\beta_t\|_2^2) $ shall be much larger than $\mathbb{E}(\|\beta_t - f \hat{\beta}_{t-1}\|_2^2)$. For example, with $f =0.9$, $\theta =0.4$, at $t = \tau +1$, $\mathbb{E}(\|\beta_t\|_2^2)$ is $0.514 \underset {i \in \Delta_{\tau}}{\Sigma} B_{i,i}  \sigma_i^2$ while $\mathbb{E}(\|\beta_t - f \hat{\beta}_{t-1}\|_2^2)$ is $0.19 \underset {i \in \Delta_{\tau}}{\Sigma} B_{i,i} \sigma_i^2$; and the ratio of $\mathbb{E}(\|\beta_t\|_2^2)$ and $\mathbb{E}(\|\beta_t - f \hat{\beta}_{t-1}\|_2^2)$ keeps increasing over time.
Recall that $\beta_t$ and $\beta_t - f \hat{\beta}_{t-1}$ are the ``noise" in (\ref{cs}) and (\ref{csres}), therefore, assuming we have an accurate estimate of principal basis at time $\tau-1$, (\ref{csres}) shall give more accurate estimate of $\hat{S}_t$ than (\ref{cs}). We show a plot of the expectations of $\| \beta_t \|_2^2$ and $\|\beta_t - f \hat{\beta}_{t-1}\|_2^2$ in Fig. \ref{plot3}.

In (\ref{cs}) and (\ref{csres}), we need an appropriate parameter $\epsilon$ which should be proportional to the ``noise" term $\|\beta_t\|_2^2$ in (\ref{cs}) or $\|\beta_t - f \hat{\beta}_{t-1}\|_2^2$ in (\ref{csres}). The ``noise" $\|\beta_t\|_2^2$ and $\|\beta_t - f \hat{\beta}_{t-1}\|_2^2$ also changes over time. Thus, we shall set the $\epsilon$ adaptively. In our work, we use  $\epsilon$ proportionally to $2\|\hat{\beta}_{t-1} \|_2^2$ for (\ref{cs}) and $\epsilon$ proportional to $2 \|\hat{\beta}_{t-1} - f \hat{\beta}_{t-2}\|_2^2$ for (\ref{csres}).

If the constraint is too tight ($\epsilon$ is too small), (\ref{cs}) or (\ref{csres}) may give solutions with some small nonzero values outside the true support set $T_t$ (also verified by our numerical experiments). As first done in \cite{dantzig}, we can also do support thresholding followed by least square estimation to reduce these errors, i.e., we can solve
\begin{eqnarray}
\hat{T}_t &=& \{i: \ (\hat{S}_t )_i \geq \gamma\} \label{supp} \\
(\hat{S}_t)_{\hat{T}_t} &=& ((\hat{P}_{t,\perp}^T)_{\hat{T}_t})^{\dag} (y_t- f \hat{P}_{t,\perp}^T \hat{L}_{t-1}) \label{ls} \\
(\hat{S}_t)_{\hat{T}_t^{c}} &=& 0  \nonumber  \\
\hat{L}_t &=& M_t - \hat{S}_t \label{Lestimate}
\end{eqnarray}

\subsection{Using model on $S_t$}
Currently, we do not use the model on $S_t$. A simple way to use it is to do modified-CS \cite{modcs} as
\begin{equation}
\min_{s} \|s_{T_\pred^c}\|_1 \ s.t. \ \| \hat{P}_{t,\perp}^T (M_t - s - f \hat{L}_{t-1})\|_2^2 \leq \epsilon  \label{modcs}
\end{equation}
with $T_{\pred}$ being an estimate of the support of $S_t$. As \cite{modcs} shows, if $T_{\pred}$ is an approximately correct estimate of currently support, (\ref{modcs}) should improve the performance of (\ref{csres}).

In previous work \cite{modcsjp,modcs_isit}, we used the previous support estimate, $\hat{T}_{t-1}$, as $T_{\text{pred}}$. This is sufficient for the problems considered in \cite{modcsjp}, where support changes very slowly over time, e.g. in case of wavelet coefficients of a medical image sequences. But for our current problem, even with one or two pixel motion between frames, the support change will be significant and $\hat{T}_{t-1}$ will have large error w.r.t. $T_t$. A better solution, is to use the motion model to predict the object(s)' location in the next frame and use this prediction to obtain $T_{\text{pred}}$ at time $t$. The details of how to do this, especially for multiple objects, will be worked out in future work.

\subsection{RR-PCP: Recursively estimating the low rank part}\label{update_P}

When some new principal directions appear, we need to detect these directions timely before the ``noise" gets large. Now, $\mathbb{E} (\|\beta_t - f\hat{\beta}\|_2^2)$ in (\ref{eqn_beta}) seems not increase with time. But this assumes $\hat{L}_{t-1} = L_{t-1}$ which is not true. When some existing directions vanish, they also need to be removed from $\hat{P}_t$. Otherwise, the number of estimated principal directions keeps increasing and thus the number of columns in $\hat{P}_{t,\perp}$, which is the number of measurements for (\ref{csres}), keeps decreasing.

At initial time, we have the training data $L^0 := [L_1, \ \cdots, \ L_{t_0}]$, which contains no sparse component.
According to our signal model (\ref{signal_model}), the data sequence $L_t$ is time correlated. Thus, we need a long sequence's data to get an accurate estimate of it's covariance. But notice in our model, the sequence $L_t - f L_{t-1}$ is time independent and has same eigenvectors as $L_t$.
Thus, we estimate principal directions of $L^0$ by estimating the covariance of $L_t - f L_{t-1}$ and computing its EVD.
Let $P_0$ and $G_0$ be the eigenvectors and (non-zero) eigenvalues of the covariance matrix of $L_t - f L_{t-1}, \ t =2, \cdots t_0$, i.e. $P_0 G_0 P_0^T = L^0 (L^0)^T$. Let $\hat{P}_{\stable}$ be an orthonormal matrix whose columns are the correctly estimated eigenvectors and let $\hat{G}_{\stable}$ be a diagonal matrix whose diagonal elements are the correspondingly correctly estimated eigenvalues.
Let $\hat{P}_t = \hat{P}_{\stable} = P_{0}$ and let $\hat{G}_t = \hat{G}_{\stable} = G_{0}$.

Our PCs update procedure is designed to estimate the current principal directions for data generated according to the piecewise stationary model on $x_t$ (and hence on $L_t$) that was described in Sec. \ref{model_L}\footnote{In future work, we will analyze real data and study existing literature to come up with more realistic models and the corresponding PCs update algorithms. For example, in practice $U$ may not be fixed, but may also rotate gradually over time.}. Assume that every $d$ frames, $k$ new directions get added or removed or both from the PCs' subspace. The newly added directions' variance starts at a small value and slowly stabilizes to the stable value. For deleted directions, we set $(\nu_t)_i = 0$ immediately and we replace $f$ by $f_d < f$ (ensures quicker decay).

Consider a change time, $t = \tau$. Let $P_{\new} := (U)_{N_\tau \setminus N_{\tau-1}}$ be the matrix containing the $k$ newly added directions. Our PCs update algorithm assumes the following,
\begin{itemize}
\item [1)]The previous additions are detected and correctly estimated before a new set gets added.
\item [2)]Let the data matrix $D$ contain $\tau_d$ frames of $\hat{L}_t - f \hat{L}_{t-1}$ after the new directions have been added. Then $\text{span}(P_{\new})$ is contained in the span of the data, $\text{span}(D)$.
\end{itemize}
Assumption 1) holds approximately if $d$ is large enough. Assumption 2) holds with high probability if $\tau_d >> k$.

\begin{algorithm}
\caption{Updating $\hat{P}_t$}
\label{algo2}

1) \textbf{Detect new appearing directions}

\begin{itemize}
\item [a)]If $\status = \stable$, compute $\|\hat{\beta}_{t-1}\|_2^2 := \|\hat{P}_{t-1,\perp}^T L_{t-1}\|_2^2$. If $\|\hat{\beta}_{t-1}\|_2^2 > \delta$, set $\status \leftarrow \detection$ and store data in D, i.e., $D \leftarrow [D, \ \hat{L}_{t-1} - f \hat{L}_{t-2}]$. If not, keep $\status = \stable$. go to step 2).
\item [b)]If $\status = \detection$,
       \begin{itemize}
       \item If there are less than $\tau_d$ frames in $D$, keep storing data difference in $D$, i.e., $D = [D, \ \hat{L}_{t-1} - f \hat{L}_{t-2}]$.
       \item If there are $\tau_d$ data frames in $D$, compute $K = (I-\hat{P}_{\stable} \hat{P}_{\stable}^T)D$.
          \begin{itemize}
          \item Estimate $\hat{P}_{\new}$ by computing the EVD of $KK^T$, i.e,
       \begin{eqnarray*}
       &&\frac{1}{\tau_d} K K^T \overset{EVD}{=} P G P^T \\
       &&T_d = \{i, \ G_{i,i} > \xi_d\} \\
       &&\hat{P}_{\new} = P_{T_d}, \ \hat{G}_{\new} = G_{T_d}
       \end{eqnarray*}
       where $G$ is a square matrix and $G_{T_d}$ is a submatrix of $G$ consisting the rows and columns indexed by $T_d$.
         \item Let $D = \emptyset$.
         \item If $\hat{P}_{\new} = \emptyset$, set $\status \leftarrow \stable$ and set $l=0$. If $\hat{P}_{\new} \neq \emptyset$, set $\status \leftarrow \rotation$ and set $l = \tau_d$.
         \end{itemize}
         \item Go to step 2)
       \end{itemize}
\item [c)]If $\status = \rotation$,
       \begin{itemize}
       \item If there are less than $\tau_r$ frames in $D$, keep storing data difference in $D$, i.e., $D = [D, \ \hat{L}_{t-1} - f \hat{L}_{t-2}]$.
       \item If there are $\tau_r$ data frames in $D$, let $K = \hat{P}_{\new}^T D$.
          \begin{itemize}
          \item  Rotate $\hat{P}_{\new}$ and $\hat{G}_{\new}$ using K, i.e.,
              \begin{eqnarray*}
              &&\frac{1}{l+\tau_r} ( l \hat{G}_{\new}  + K K^T) \overset{EVD}{=}P G P^T \\
              && T_r = \{i: \ G_{i,i} > \xi_r\} \\
              && \hat{P}_{\new} = (\hat{P}_{\new} P)_{T_r}, \ \hat{G}_{\new} =(G )_{T_r}
              \end{eqnarray*}
          If $P$ is approximately an identity matrix, let $\hat{P}_{\stable} \longleftarrow [\hat{P}_{\stable}, \hat{P}_{\new}]$, $\hat{G}_{\stable} \longleftarrow [\hat{G}_{\stable}, \hat{G}_{\new}]$. Set $\status \leftarrow \stable$, and let $\hat{P}_{\new} = \emptyset$, $\hat{G}_{\new} = \emptyset$, $l=0$. If not keep $\status \longleftarrow \rotation$ and let $l = l +\tau_d$.
          \item $D = \emptyset$.
          \end{itemize}
       \item Go to step 2).
       \end{itemize}
\end{itemize}

2) \textbf{Remove decayed directions from $\hat{P}_{\stable}$.}
 \begin{itemize}
 \item If there are less than $\tau_{\del}$ data in $D_{\del}$, keep store data difference in $D_{\del}$, i.e., $D_{\del} = [D_{\del}, \ \hat{L}_{t-1} - f \hat{L}_{t-2}]$.
 \item If there $\tau_{\del}$ data in $D_{\del}$, detect decayed directions as follows
          \begin{itemize}
          \item Find $T_{\del} :=  \{i:\  \frac{1}{\tau_{\del}}(\hat{P}_{\stable})_i^T D_{\del} D_{\del}^T (\hat{P}_{\stable})_i < 0.05 (\hat{G}_{\stable})_{i,i} \}$.
          \item Remove $(\hat{P}_{\stable})_{T_{\del}}$ from $\hat{P}_{\stable}$ and remove $(\hat{G}_{\stable})_{T_{\del}}$ from $\hat{G}_{\stable}$.
          \item Set $D_{\del} = \emptyset$.
          \end{itemize}
 \end{itemize}

3) \textbf{Let $\hat{P}_t = [\hat{P}_{\stable}, \hat{P}_{\new}]$.}
\end{algorithm}

We split the estimate of PCs' basis, $\hat{P}_t$, into two parts, $\hat{P}_t = [\hat{P}_{\stable}, \  \hat{P}_{\new}]$ where $\hat{P}_{\stable}$ is the ``stable" (correctly estimated) set of principal directions and $\hat{P}_{\new}$ are the new ones which are still being rotated and corrected. We would like to compute an initial estimate of $P_{\new}$ as soon as possible (using only a few frames after $\|\hat{\beta}_t\|^2$ exceeds a threshold). Say we use $\tau_d$ frames and let the matrix $D$ contains $\hat{L}_{t-1} - f \hat{L}_{t-1}$ for these frames. We can compute an initial estimate of the new directions, $\hat{P}_{\new}$, by computing the principal directions of the sample covariance matrix of $(I - \hat{P}_{\stable} \hat{P}_{\stable})^T D$. This is done by step 1.b) of Algorithm \ref{algo2}.
By assumption 2), if $\tau_d > k$, then we would have found the correct span, i.e. $\text{span}(\hat{P}_{\new}) \supseteqq \text{span}(P_{\new})$. But notice that without enough data, even though $\text{span}(\hat{P}_{\new})$ contains $\text{span}(P_{\new})$, it will typically contain many extra directions. As more data comes in, we keep rotating $\hat{P}_{\new}$ every-so-often until variances along some directions become approximately zero and these get thresholded out. Once this has happened, the estimated rotation matrix $P$ along the existing directions becomes close to identity and remains this way. This is the time we can add $\hat{P}_{\new}$ into $\hat{P}_{\stable}$. This is done by step 1.c) of Algorithm \ref{algo2}.

When the variances along some directions in $\hat{P}_{\stable}$ begin to decrease and eventually decay to zero, we compute the variance of last $\tau_{\del}$ frames along $\hat{P}_{\stable}$ and then remove directions with small variance from $\hat{P}_{\stable}$. This is done by step 2) of Algorithm \ref{algo2}.

The above PCs update procedure is summarized in Algorithm \ref{algo2}. In Algorithm \ref{algo2}, $D$ and $D_{\del}$ are data matrix to store the data difference $\hat{L}_t - f \hat{L}_{t-1}$. The parameters, $\tau_d$, $\tau_r$, and $\tau_{\del}$, are the length of each data piece we use to detect new directions, to rotate and correct newly added directions, and to remove decayed directions, respectively. We use two small thresholds, $\xi_d$ and $\xi_r$, to detect new directions $\hat{P}_{\new}$ in step 1b) and to threshold extra directions out from $\hat{P}_{\new}$ in step 1c). They are proportional to the total variance along all existing stable directions.

\subsection{A complete algorithm}
The complete algorithm of real-time robust PCP is given in Algorithm \ref{algorpcp}.

In Algorithm \ref{algorpcp}, we compute $\hat{P}_{t,\perp}$, orthogonal complement of $\hat{P}_t$ or equivalently the null space basis of $\hat{P}_t^T$, using the QR decomposition of $\hat{P}_t$ \cite{matrix_analysis}. Suppose $\hat{P}_t$ is $m \times r$ matrix with $m>>r$. We find an $m \times m$ orthonormal matrix $H$ such that
\begin{eqnarray}
\hat{P}_t &\overset{QR}{=}& H J = \begin{bmatrix} H_1 \ H_2 \end{bmatrix} \begin{bmatrix} J_1 \\ 0 \end{bmatrix} \nonumber
\end{eqnarray}
where $J_1$ is an $r \times r$ upper triangular matrix. $H_1$ consists of the first $r$ columns of $H$ and $H_2$ is made up of the last $m-r$ columns. The columns of $H_2$ span the null space of $\hat{P}_t^T$ and we let $\hat{P}_{t,\perp} = H_2$.

\begin{algorithm}
\caption{Real-time Robust PCP (noise canceled) }
\label{algorpcp}

Training:
Given training data $L_0 = [L_1, \cdots, L_{t_0}]$, estimate principal components of $L_0$ by computing the eigen-pairs of the sample covariance of $L_t - f L_{t-1}$. Let $P_0$ and $G_0$ denote the eigenvectors and eigenvalues. Set $\hat{P}_{\stable} = P_0$, $\hat{G}_{\stable} = G_0$.

At time $t=t_0$,
\begin{itemize}
\item Set $\status = \stable$. Let $\hat{P}_t =\hat{P}_{\stable}$, $\hat{G}_t = \hat{G}_{\stable}$, $\hat{P}_{\new} =\emptyset$, $\hat{G}_{\new} = \emptyset$.
\item Let $D = \emptyset$, $l = 0$, $D_{\del} = \emptyset$. 
\end{itemize}

FOR $t>t_0$, do the following:
\begin{itemize}
\item [1)] Estimate PCs' subspace of low rank part using Algorithm \ref{algo2} and compute $\hat{P}_{t,\perp}$.

\item [2)] Estimate sparse part, $S_t$, by solving (\ref{csres}) with $\epsilon = 2 \|\hat{\beta}_{t-1} - f \hat{\beta}_{t-2}\|_2^2$.

\item [3)] Support thresholding and least square estimation: do (\ref{supp}), (\ref{ls}), and (\ref{Lestimate}).

\item [4)] Increment $t$ by $1$ and go to step 1).

\end{itemize}

\end{algorithm}

\section{Experiment results}\label{result}

We simulated $L_t \in \mathbb{R}^{128 \times 1}$ using the model described in Sec.\ref{model_L}.
The first $t_0 = 5\times 10^3$ frames contains no sparse part and we use it as training data.
The sparse vector, $x_t$, follows a AR-1 model with parameter $f=0.9$. There are $32$ principal directions with variances ranging from $1\times 10^4$ to $9$.
Recall that in our model, $L_t$ is time correlated and the sequence $L_t - f L_{t-1}$ has same eigenvectors as $L_t$, we get initial estimate of PCs' subspace by estimating the covariance of $L_t - f L_{t-1}$ and computing its EVD.

The sparse component $S_t \in \mathbb{R}^{128 \times 1}$ first arises at time $t = t_0 + 1$. The nonzero entries of $S_t$ has positive magnitude $5$, which is usually much smaller than magnitude of the nonzero entries of $x_t$. For $t>t_0+1$, the support of $S_t$ changes following the model described in Sec.\ref{model_S}, resulting in a low rank matrix $S$. 

\begin{figure*}
\centerline{
\psfig{file =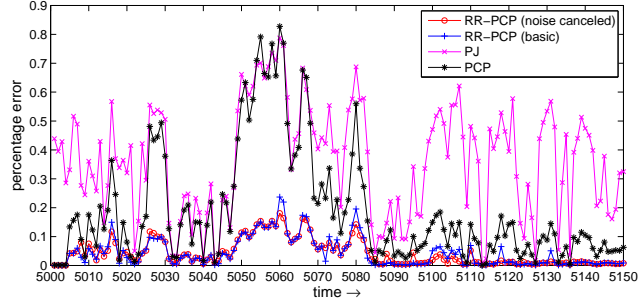,height = 4.5cm}
}
\caption{Comparison of RR-PCP (noise canceled), RR-PCP (basic), PJ and PCP}\label{plot1}
\end{figure*}

\begin{figure*}
\centerline{
\psfig{file = 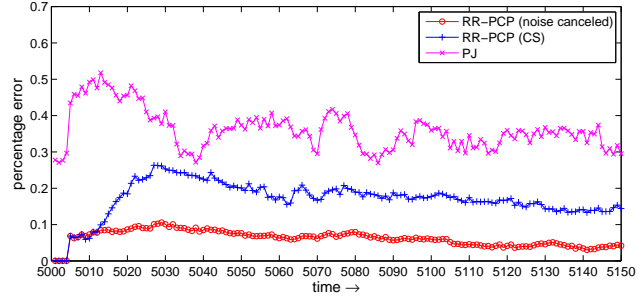,height = 4.5cm}
}
\caption{Monte Carlo averaging for three on-line methods, RR-PCP (noise canceled), RR-PCP (basic), and PJ}\label{plot2}
\end{figure*}
At time $t = t_0 + 5$, we add one new direction $P_{\new}$ with variance $50$ to PCs' basis and let it starts at a small value with $\theta = 0.4$. It slowly stabilizes to the variance $50$.
At time $t = t_0 + 100$, one existing direction (not $P_{\new}$) begins to decay with $f_d = 0.1$.

We do RR-PCP (noise canceled), RR-PCP (basic), PJ and PCP using the data generated as described above and plot the percentage error in Fig. \ref{plot1}. The percentage error is defined as
\begin{equation*}
\text{percentage error}:= \frac{\|S_t-\hat{S}_t\|_2}{\|S_t\|_2}
\end{equation*}
In Fig. \ref{plot1}, $S_{1:t}$ is $1024 \times t$ dimensional at time $t$, but its rank ranges from $0$ to $51$.

For RR-PCP (noise canceled), we do algorithm \ref{algorpcp} with $\tau_d = \tau_r =\tau_{\del} = 20$.
At $t= t_0+6$, it detects the appearance of new directions and set $\status = \detection$.
At $t= t_0 + 26$, a new piece of data containing $\tau_d$ frames are available, RR-PCP do step 1b) of Algorithm \ref{algo2} and get $\hat{P}_{\new}$, an estimate of the new direction $P_{\new}$. There are $7$ new directions in $\hat{P}_{\new}$, and the coherence between these new estimated directions and the true one $P_{\new}$ ranges from $0.9393$ to $0.0051$. So, with $\tau_d$ frames of data, it approximately finds a subspace containing $P_{\new}$ and some extra directions.
For $t>t_0+26$, we do step 1c) of Algorithm \ref{algo2} to rotate $\hat{P}_{\new}$ closer to the true one and threshold out those extra directions for every $\tau_r = 20$ new frames of data.
For example, at $t=t_0 + 47$ when a new piece of data is available, we do step 1c) which rotates $\hat{P}_{\new}$ closer to the true $P_{\new}$ and get 2 directions thresholded out. The maximum coherence of $\hat{P}_{\new}$ and $P_{\new}$ goes up to $0.9505$. At time $t =t_0 + 68$, another two directions are thresholded out and the maximum coherence of $\hat{P}_{\new}$ and $P_{\new}$ goes up to $0.9526$; at time $t=t_0+110$, the rotation matrix $P$ is close to an identity matrix (on-diagonal elements larger than $0.9999$ and off-diagonal elements smaller than $0.01$). Only one direction is left in $\hat{P}_{\new}$, with coherence $0.9553$ to $P_{\new}$. It sets $\status = \stable$ and adds $\hat{P}_{\new}$ to the stable set of principal directions.
At time $t = t_0 +126$, it removes the deleted direction from the estimated PCs' basis successfully.

\begin{figure}
\centerline{
\psfig{file = 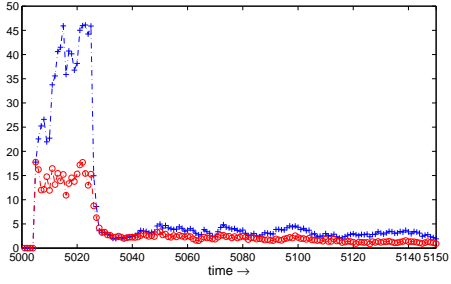, width = 7cm}
}
\caption{$\mathbb{E}\|\beta_{t}\|_2^2$ v.s. $\mathbb{E}\|\beta_t - f \hat{\beta}_{t-1}\|_2^2$}\label{plot3}
\end{figure}

For RR-PCP (basic), we do same thing as RR-PCP (noise cancelled) but replace (\ref{csres}) with (\ref{cs}) and replace (\ref{ls}) by doing LS on $y_t$. We see that error of RR-PCP (basic) is larger than RR-PCP (noise cancelled) because it does not use the information contained in $\hat{L}_{t-1}$, i.e. $\|\beta_{t}\|^2$ is larger than $\|\beta_t - f \hat{\beta}_{t-1}\|^2$(see Fig.\ref{plot3}).

For PJ, we solve (\ref{PJ}) with $A = [\hat{P}_t, \ \hat{P}_{t,\perp}, \ I]$. PJ recovers $x_t$ and $S_t$ while RR-PCP (noise cancelled) and RR-PCP (basic) cancel the term $x_t$ by $\hat{P}_{t,\perp}^T$.
Recall that $x_t$ has variance ranging from $1\times 10^4$ to $9$, the magnitude of $x_t$ is much larger than $S_t$. PJ recovers the significant part $x_t$ and cannot get $S_t$ recovered correctly.

For the off-line method PCP, at each time $t$, we solve (\ref{PCP}) using all available data frames\footnote{We use Accelerated Proximal Gradient method with code available at \url{http://perception.csl.illinois.edu/matrix-rank/sample_code.html}.}, $[M_1, \ \cdots, \ M_t]$, and plot the error for current frame $S_t$. The error of PCP is large because the support of $S_t$ is time correlated and $S_t$ does not has random signs. To implement PCP in a causal fashion, it requires about $200$ - $300$ seconds at every time $t$, while RR-PCP takes about $1.7$ seconds at every time $t$.

We do $50$ times Monte Carlo simulation for three on-line methods, average the percentage error and plot them in Fig. \ref{plot2}. As can be seen from Fig.\ref{plot2}, our method RR-PCP (noise canceled), gives the smallest error.
In Fig.\ref{plot3}, we plot the expectation of $\|\beta_{t}\|_2^2:=\|\hat{P}_{t,\perp}^T L_t\|_2^2$ and $\|\beta_t - f \hat{\beta}_{t-1}\|_2^2:= \|\hat{P}_{t,\perp}^T (L_t - f \hat{L}_{t-1})\|_2^2$ for RR-PCP (noise canceled). It shows that, for $t<t_0 +26$ when there are some missing principal directions, $\|\beta_t - f \hat{\beta}_{t-1}\|_2^2$ is much smaller than $\|\beta_{t}\|_2^2$. For $t>t_0 +26$, RR-PCP (noise canceled) gets an estimate $\hat{P}_{\new}$ and adds it to the estimate of current PCs' basis, thus, both $\|\beta_{t}\|_2^2$ and $\|\beta_t - f \hat{\beta}_{t-1}\|_2^2$ decreases. However, $\|\beta_t - f \hat{\beta}_{t-1}\|_2^2$ is still slightly less than $\|\beta_{t}\|_2^2$ due to the time correlated model on $L_t$.
Recall that $\|\beta_t\|_2^2$ is the noise in (\ref{cs}) and $\|\beta_t - f \hat{\beta}_{t-1}\|_2^2$ is the noise in (\ref{csres}), that is the reason why RR-PCP (noise canceled) is better than RR-PCP (basic).
\section{Discussion and future work}

In this work, we used a simple motion model on the sparse vector $S_t$ as explained in Sec. \ref{model_S}. Under this model, the support of $S_t$ changes slowly over time, resulting in a low rank matrix $S$. Because of this, PCP is unable to distinguish $S$ from the low rank $L$. But our method, RR-PCP, works because it does not require the sparse matrix $S$ to to be uniformly random. In this work, we have not utilized the correlated support change of $S_t$ to our advantage. But, in fact, RR-PCP can be improved significantly by using this knowledge and by adapting the modified-CS idea of \cite{modcs,modcsjp} to incorporate motion prediction. We can use the knowledge of the object's motion model and the previous support estimate to obtain the current support prediction $T_{\pred}$ of the sparse part. If this prediction is accurate and is used in (\ref{modcs}), with an appropriately chosen $\epsilon$, the reconstruction error should reduce significantly, especially when the support size of $S_t$ is large. In future work, we will develop realistic motion models and corresponding motion prediction algorithms to get reliable support predictions of the sparse part. We will also analyze their performance, first assuming $P_t$ is perfectly known and later for the practical case of $P_t$ unknown.

Our PCs updating procedure is designed for the data generated according to the piecewise stationary model on $x_t$ while $U$ is a constant but unknown orthonormal matrix. In future work, we will analyze real data and study existing literature to come up with more realistic models and the corresponding PC update algorithms.

For very-large scale data, it is computationally and memory intensive to compute $\hat{P}_{t,\perp}$. In future work, we will develop computational efficient alternatives. For example we can use the fact that $\|\hat{P}_{t,\perp}^T z \|_2 = \|\hat{P}_{t,\perp} \hat{P}_{t,\perp}^T z \|_2  = \|(I - \hat{P}_{t}\hat{P}_{t}^T) z \|_2$.  A somewhat related work is Jin-Rao's approach \cite{rpca_regression} which solves
\begin{equation}
\min_{\alpha,s} ||s||_1 \ s.t. \ || M_t - P_t \alpha - s||_2^2 \le \epsilon  \label{Jin-Rao}
\end{equation}
In \cite{rpca_regression}, the matrix $P_t$ is a known and fixed regression coefficients' matrix, which is no longer true in our problem. We can use the time correlated model on $x_t$ (and hence on $\alpha_t$) and the motiom model on $S_t$ to modify (\ref{Jin-Rao}) following a similar way of RR-PCP.

\bibliographystyle{ieeepes}
\bibliography{tipnewpfmt}

\begin{thebibliography}{10}

\bibitem{rpca_1}
P.~Huber,
\newblock {\em Robust statistics},
\newblock Wiley and Sons, 1981.

\bibitem{rpca_caramanis}
Huan Xu, C.~Caramanis, and S.~Mannor,
\newblock ``High dimensional principal component analysis with contaminated
  data'',
\newblock in {\em Networking and Information Theory, 2009. ITW 2009. IEEE
  Information Theory Workshop on}, jun. 2009, pp. 246 --250.

\bibitem{Roweis98emalgorithms}
Sam Roweis,
\newblock ``Em algorithms for pca and spca'',
\newblock in {\em in Advances in Neural Information Processing Systems}. 1998,
  pp. 626--632, MIT Press.

\bibitem{rpca_2}
F.~De~La Torre and M.~Black,
\newblock ``A framework for robust subspace learning'',
\newblock in {\em International Journal on Computer vision}, 2003, pp.
  54:117--142.

\bibitem{sequentialSVD}
Matthew Brand,
\newblock ``Incremental singular value decomposition of uncertain data with
  missing values'',
\newblock in {\em In ECCV}, 2002, pp. 707--720.

\bibitem{Li03anintegrated}
Yongmin Li, Li~qun Xu, Jason Morphett, and Richard Jacobs,
\newblock ``An integrated algorithm of incremental and robust pca'',
\newblock in {\em Proceedings of IEEE International Conference on Image
  Processing}, 2003, pp. 245--248.

\bibitem{ipca_5}
Lei Xu and A.~Yuille,
\newblock ``Robust principal component analysis by self-organizing rules based
  on statistical physics approach'',
\newblock in {\em IEEE Trans. Neural Networks}, 1995, pp. 131--143.

\bibitem{ipca_weightedand}
D.~Skocaj and A.~Leonardis,
\newblock ``Weighted and robust incremental method for subspace learning'',
\newblock oct. 2003, pp. 1494 --1501 vol.2.

\bibitem{Torre01robustprincipal}
F.~De~la Torre and M.J. Black,
\newblock ``Robust principal component analysis for computer vision'',
\newblock 2001, vol.~1, pp. 362 --369 vol.1.

\bibitem{rpca}
Emmanuel~J. Cand{\`e}s, Xiaodong Li, Yi~Ma, and John Wright,
\newblock ``Robust principal component analysis?'',
\newblock {\em Submitted for publication}.

\bibitem{rpca_Chandrasekaran}
Venkat Chandrasekaran, Sujay Sanghavi, Pablo~A. Parrilo, and Alan~S. Willsky,
\newblock ``Sparse and low-rank matrix decompositions'',
\newblock in {\em Allerton'09}, 2009.

\bibitem{error_correction_PCP}
A.~{Ganesh}, J.~{Wright}, X.~{Li}, E.~J. {Candes}, and Y.~{Ma},
\newblock ``{Dense Error Correction for Low-Rank Matrices via Principal
  Component Pursuit}'',
\newblock {\em ArXiv e-prints}, January 2010.

\bibitem{decodinglp}
E.~Candes and T.~Tao,
\newblock ``Decoding by linear programming'',
\newblock {\em IEEE Trans. Info. Th.}, vol. 51(12), pp. 4203 -- 4215, Dec.
  2005.

\bibitem{rpca_regression}
Yuzhe Jin and Bhaskar Rao,
\newblock ``Algorithms for robust linear regression by exploiting the
  connection to sparse signal recovery'',
\newblock in {\em IEEE Intl. Conf. Acous. Speech. Sig.Proc.(ICASSP)}, 2010.

\bibitem{rpca_regression_sparse}
Kaushik Mitra, Ashok Veeraraghavan, and Rama Chellappa,
\newblock ``A robust regression using sparse learing for high dimensional
  parameter estimation problems'',
\newblock in {\em IEEE Intl. Conf. Acous. Speech. Sig.Proc.(ICASSP)}, 2010.

\bibitem{error_correction_PCP_l1}
John Wright and Yi~Ma,
\newblock ``Dense error correction via l1-minimization'',
\newblock {\em IEEE Transactions on Information Theory}, 2009.

\bibitem{Laska_exactsignal}
J.N. Laska, M.A. Davenport, and R.G. Baraniuk,
\newblock ``Exact signal recovery from sparsely corrupted measurements through
  the pursuit of justice'',
\newblock in {\em Signals, Systems and Computers, 2009 Conference Record of the
  Forty-Third Asilomar Conference on}, Nov 2009, pp. 1556 --1560.

\bibitem{dantzig}
E.~Candes and T.~Tao,
\newblock ``The dantzig selector: statistical estimation when p is much larger
  than n'',
\newblock {\em Annals of Statistics}, 2006.

\bibitem{modcs}
Wei Lu and Namrata Vaswani,
\newblock ``Modified basis pursuit denoising (modified-bpdn) for noisy
  compressive sensing with partially known support'',
\newblock in {\em IEEE Intl. Conf. Acous. Speech. Sig.Proc.(ICASSP)}, 2010.

\bibitem{modcsjp}
N.~Vaswani and W.~Lu,
\newblock ``Modified-cs: Modifying compressive sensing for problems with
  partially known support'',
\newblock {\em IEEE Trans. Signal Processing}, September 2010.

\bibitem{modcs_isit}
N.~Vaswani and Wei Lu,
\newblock ``Modified-cs: Modifying compressive sensing for problems with
  partially known support'',
\newblock in {\em ISIT 2009}, June 2009, pp. 488 --492.

\bibitem{matrix_analysis}
R.~A. Horn and C.~R. Johnson,
\newblock {\em Matrix Analysis},
\newblock Cambridge University Press, February 1990.

\end{thebibliography}

\end{document}